\pdfoutput=1
\documentclass[conference, hidelinks, final]{IEEEtran}

\usepackage[english]{babel}
\usepackage[T1]{fontenc}
\usepackage{graphicx}
\usepackage{amsmath}
\usepackage{amsthm}
\usepackage{amssymb}
\PassOptionsToPackage{hyphens}{url} 
\usepackage{hyperref}
\usepackage{fancyhdr}
\usepackage{listings}
\usepackage{color}
\usepackage{xcolor}
\usepackage{xfrac}
\usepackage{cite}

\usepackage[
	protrusion=true, activate={true,nocompatibility},final, 
	tracking=true,kerning=true,spacing=true,factor=1100]{microtype} 
\usepackage[subtle]{savetrees} 

\graphicspath{ {images/} }

\title{Leveraging the HW/SW Optimizations and Ecosystems that Drive the AI Revolution\vspace{-5pt}}

\author{
	\IEEEauthorblockN{
		Humberto Carvalho\IEEEauthorrefmark{1},
		Pavel Zaykov\IEEEauthorrefmark{1},
		and Asim Ukaye\IEEEauthorrefmark{2}
	}
	\IEEEauthorblockA{
		\IEEEauthorrefmark{1} Honeywell International s.r.o., Advanced Technology Europe, Brno, Czechia\\ 
		\IEEEauthorrefmark{2} Honeywell Technology Solutions, Flight Systems COE, Bengaluru, India  \\
		humberto.carvalho@honeywell.com, pavel.zaykov@honeywell.com, asim.ukaye@honeywell.com
	}

\vspace{-20pt}
}

\begin{document} 
\maketitle
\begin{abstract}   
    This paper presents a state-of-the-art overview on how to architect, design, and optimize Deep Neural Networks~(DNNs) such that performance is improved and accuracy is preserved. 
	The paper covers a set of optimizations that span the entire Machine Learning processing pipeline. 
	We introduce two types of optimizations. The first alters the DNN model and requires NN re-training, while the second does not.
	We focus on GPU optimizations, but we believe the presented techniques can be used with other AI inference platforms. 
    To demonstrate the DNN model optimizations, we improve one of the most advanced deep network architectures for optical flow, RAFT~\cite{paper:RAFT}, on a popular edge AI inference platform (Nvidia Jetson AGX Xavier).
\end{abstract}

\section{Introduction}
\label{sec:intro}

Artificial Intelligence (AI), and in particular Machine Learning (ML) algorithms, have shown tremendous breakthroughs in many areas of research. As an example, the air and ground transportation industries are undergoing one of the most significant transitions from pilot/driver-centric to fully-autonomous operation. To make the transition, the on-board computing devices have to process data from multitude of sensors and perceive the surrounding environment. This area of research is known as machine perception, where the on-board computing algorithms perform a wide-range of tasks such as image classification, image segmentation, object detection, and depth estimation. Besides the transportation industry, DNNs are applied in a wide range of subjects such as language translation, 3D medical research, and speech recognition. 

A common reference point of the recent ML revolution~\cite{paper:history_began_from_alexnet} is the remarkable accuracy of Alexnet~\cite{paper:alexnet} at the ImageNet Large Scale Visual Recognition challenge~\cite{paper:imagenet} in 2012. 
Alexnet scored a Top-5 error rate of 15.3\%, more than 10\% ahead of the second best. 
Compared to its rivals at the competition, Alexnet was based on Deep Neural Networks~(DNNs), more specifically on a Convolutional Neural Network~(CNN). Alexnet's success was a combination of several novelties, most notably: (1) the use of distributed GPU training to accelerate computation, (2) the use of a large  dataset with over 1.4 million images, (3) the use of novel DNN architectural features such as the ReLU activation function that facilitates back-propagation.

State-of-the-art DNNs achieve notable gains in accuracy through novel architectures and the availability of exceptionally high computational resources that are required to explore, train, and execute these DNNs. There is a tight correlation between the accuracy of a DNN model and its computational complexity, where deeper/wider models with higher compute requirements achieve better accuracy. Since Alexnet, the computational requirements of DNN models have increased exponentially. For example, a modern transformer machine vision model such as ViT-G/14~\cite{paper:vitg14} requires 612 GFLOPs during forward propagation. 

The success of ML systems can be attributed to several factors. The software ecosystem, which allows researchers to focus on development rather than struggling with implementations; the availability of large datasets comprised of millions of examples; and finally, the advances in computational performance of state-of-the-art accelerators like GPUs. 
 
Many ML breakthroughs can be attributed to computational leaps. As an example, the emergence of back-propagation in the 80s can be attributed to the availability of computing platforms that supported floating-point computation.  Today,  computational leaps significantly impact the development and deployment of DNN based  systems; as novel DNN architectures can be explored within a shorter period of time, and then deployed on the edge within the  tighter resource constraints of edge platforms like mobile and embedded.

In this paper, we present an overview of the SW ecosystem and the state-of-the-art DNN optimization techniques that have driven the AI revolution. We survey battle-tested optimization methods that have seen great industry adoption and can be applied to a wide range of NN architectures, inference engines, and HW platforms. We provide an overview of the edge HW and SW ecosystem (inference engines, frameworks, and compilers) which leverage said optimizations methods, analyze their architecture and compare their performance and portability trade-offs. Our analysis covers the entirety of the ML inference pipeline, such as pre-/post-processing and encoding/decoding. To provide an overview behind the motivations of the optimizations, we introduce arithmetic intensity and roofline analysis, key principles in performance optimization. 
Overall, these contributions are of great value for SW engineers looking to deploy and optimize DNN applications on the edge. We believe our contribution differentiates from other works \cite{paper:deng2020model}, \cite{paper:capra2020hardware} as it presents all this information in a pratical and concise report. 

The paper is structured as follows. Section~\ref{sec: background} surveys the AI edge landscape in terms of the computing platforms, frameworks, and standards that contribute to the AI Revolution.  Section~\ref{sec:compute_background} introduces the compute background that drives DNN performance and illustrates the motivation behind DNN optimization techniques. Sections~\ref{sec:pipeline}, \ref{sec:nn_archs}, \ref{sec:pipeline_optimizations} present the main computing stages of a modern ML processing pipeline and outline the performance bottlenecks and respective optimization methods for each pipeline stage. Finally, section~\ref{sec:experimental} demonstrates the process of DNN optimization on an advanced DNN architecture for estimating optical flow (RAFT~\cite{paper:RAFT}) using a popular edge AI inference platform (Nvidia Jetson AGX Xavier).
The optimizations boost the performance by 6x while preserving the original model architecture. 

\section{The edge AI HW \& SW Landscape}
\label{sec: background}

Given the unprecedented compute requirements of DNNs, various computer architectures consisting of a combination of CPUs, GPUs, FPGAs, and ASICs have been proposed by the industry. 
Nvidia was the first to widely commercialize high performance HW accelerators for Matrix Multiplication known as Tensor Cores~\cite{paper:tensor_cores}. For the edge market, Nvidia introduced a specialized family of designs referred as Jetson (for embedded) and Drive (for automobile). 
For embedded, Intel offers the TigerLake CPU with an integrated GPU that is capable of AI inference. Future releases are expected to have specialized HW acceleration for Matrix Multiplication. 
Xilinx, a traditional supplier of state-of-the art SRAM-based FPGAs, has also recently introduced the Versal AI Edge series
with programmable AI cores targeting  Matrix Multiplication. 

The vendor-specific HW accelerators are supported by SW ecosystems, each with a custom set of performance-optimized compilers, runtimes and optimization tools. These ecosystems provide a highly efficient and HW-accelerated set of primitives to accelerate a wide range of tasks. For ML, these primitives are known as operators, and provide the building blocks to train and deploy DNNs. Operators are hardware optimized, and often include DNN specific optimizations.  Nvidia provides the most complete ecosystem with tools such as cuBLAS, cuDNN, TensorRT and DeepStream SDK. 
Other manufacturers such as Intel, Xilinx, Texas Instruments, and ARM, also provide their own software ecosystem. 
Although these ecosystems do not share a common API, they generally share a common deployment workflow.  After training, a DNN is exported to an exchange format such as ONNX
and NNEF, that is later executed using the vendors AI inference engine.

To support the latest DNN architectures proposed in the state-of-the-art, each vendor shall support the latest operators proposed in these novel DNN architectures. 
To reduce development effort of maintaining operator compatibility and to increase the portability of emerging DNN architectures, several approaches have been proposed in the industry and academia. These approaches can be categorized into three groups, each with unique performance/portability trade-offs: (1) cross-platform APIs, (2) cross-platform Compute APIs, and (3) compiler-based solutions. An overview of each is provided in the remainder of this section. 

\emph{Cross-platfrom APIs}: To unite the performance-optimized vendor runtimes under a cross-platform API, several standards have been proposed.  The ONNX community introduced ONNX Runtime 
which supports various inference engines such as TensorRT, Intel OpenVino, and Xilinx Vitis AI. Khronos introduced the Neural Network Extensions to OpenVX and currently supports MIVisionX by AMD.  Android has introduced the Android Neural Networks API (NNAPI) runtime, which currently supports a wide-range of ARM-based devices. Finally, Microsoft recently proposed DirectML that supports AMD, Intel and Nvidia devices. DirectML is the only cross-platform API that supports DNN training.

\emph{Cross-platform Compute APIs:} The cross-platform API approach maximizes performance, but does not fully solve the portability problem as the hardware vendors must still implement a performance optimized runtime for each HW architecture. To build a cross-platform runtime, the Khronos community advocates cross-platform compute APIs like OpenCL and Vulkan. This approach trades some performance for improved portability, as DNN operators are not tailored to each hardware platform; in addition, these APIs do not provide a common interface to leverage custom hardware accelerators like Tensor Cores. Popular ML development frameworks like  Pytorch and TensorFlow have introduced deployment runtimes that leverage these APIs, such as torchscript, Tensorflow Lite, and NCNN.

\emph{Compiler-based Solutions:} To reduce the labor-intensive process of optimizing NN operators for various architectures, both the academia and industry have explored compilers to automate the  process of optimizing and deploying DNN architectures. In this approach, a compiler ingests and optimizes a DNN model, producing a binary which can then be used to execute the model on a target HW platform. By using compilers, emerging operators proposed in the state-of-the-art can be automatically tuned, reducing the optimization process from days/weeks down to minutes. The most popular compilers up to date are: Apache TVM~\cite{paper:apache_tvm}, Glow by Facebook~\cite{paper:pytorch_glow}, and XLA and Jax by Google~\cite{paper:google_jax}.

\section{Compute Background}
\label{sec:compute_background}

\subsection{GPGPU Computation}
\label{sec:gpgpu}

Graphical Processing Units (GPUs) are a specialized HW unit designed for the throughput of data-parallel problems. 

In GPUs, the majority of the chip area is dedicated to functional units managed by simple in-order and relatively slow pipelines. Modern GPUs often have thousands of functional units, which vastly outnumbers the functional units in a CPU.  While each functional unit within a GPU is relatively slow compared to CPUs, the total number of units makes GPUs orders of magnitude faster than CPUs. 

To keep the functional units busy, GPUs rely on thread oversubscription, where each aggregate of functional units manages threads in groups of a certain size. 
Whenever one of these thread-groups stalls due to cache misses or other types of stalls, the in-order pipeline is able to switch to another group very efficiently. Unlike CPUs, thread context is kept within the hardware itself, thus context switching can be performed within a clock cycle. This architectural model fits well with data-throughput problems. But in order to fully utilize a GPU, tens of thousands of threads are required to keep the functional units saturated and thus hide cache/pipeline related latency.

\subsection{Generalized Matrix Multiply (GEMM)}
\label{sec:gemm}
Matrix Multiplication is a good example of a data-parallel problem that can be highly parallelized. Signal processing applications and fluid dynamics simulations, for example, often require Matrix Multiplications. Thus, efficiently speeding up Matrix Multiplication would impact various application domains. Moreover, Generalized Matrix Multiplications~(GEMMs) are a fundamental building block for DNNs operators, such as fully-connected layers, recurrent layers, and convolution layers. 

GEMM performance optimization is an active area of exploration, where the current research is focused on leveraging the hardware architecture to its full potential~\cite{book:d2lai}. Both, the CPUs and GPUs have a set of registers, multi-level caches of increasing size but decreasing bandwidth, and DRAM memory. From a hardware design perspective, it is easier to add more computational resources than memory bandwidth. Thus, the compute units within CPUs and GPUs are capable of performing many times the number of operations than what the DRAM interface can provide. Only the lowest cache levels, those closest to the compute cores, are capable of providing enough bandwidth to saturate the functional units within them. Thus, to implement GEMMs efficiently, matrices are divided into blocks and stored in the lower cache levels during processing. 

Dividing the original Matrix into blocks creates a bin-packing problem. Depending on the Matrix dimensions and the GPU architecture characteristics, performance efficiency can be lost due to: (1) matrix dimensions that are not evenly divisible into blocks, thus resulting in an uneven load distribution among the blocks, (2) uneven load distribution within the GPU, which occurs whenever the number of blocks is not evenly divisible by the number of computing cores, (3) small work blocks without sufficient work re-use to fully utilize each GPU core.

\subsection{Arithmetic Intensity}
\label{sec:arithmetic_intensity}
Among the most common performance metrics for modern processors are the maximum theoretical throughputs, such as the peak number of math operations per second \emph{OP/s}, and the peak memory throughput in \emph{GB/s}. Using the math and memory capabilities of the processor, the arithmetic intensity is computed by $\tfrac{OP/s}{GB/s}$~\cite{site:gtc_gpu_compute}. The arithmetic intensity defines the difference between the hardware compute capabilities \emph{OP/s} and memory subsystem data-rate \emph{GB/s}.

Application performance can also be specified by using \emph{OPs} and \emph{GBs}. For example, matrix multiplication can be categorized in terms of the required math operations and memory accesses per second. 
Thus,  arithmetic intensity can also be used to analyze algorithms and understand whether they are either compute- or memory-bound. An algorithm is memory-bound if its arithmetic intensity is lower than the hardware's, and compute-bound otherwise. There are not many algorithms that are compute-bound, however, matrix multiplication is an example of a such algorithm.

Arithmetic intensity only counts algorithmic operations with the assumption that they fully saturate the underlying resources, and thus, is only a first order approximation; some profilers, such as Nvidia NSight Compute provide accurate system-based metrics that take all operations into account.

\subsection{Roofline Analysis}
Roofline analysis~\cite{site:gtc_roofline} is based on arithmetic analysis. It quantifies the maximum theoretical performance that a given hardware platform can provide for a given algorithm and input. Given the highly parallel nature of a GPU, the overall execution time is defined by Equation~\ref{equation:roofline_time}, that is, the time period required to perform the necessary \emph{OPs}, or the time period necessary to move the required data \emph{GBs}, whichever is greater. 
\begin{equation}
\small
	\label{equation:roofline_time}
	Time = max
	\begin{cases}
		sw^{OPs} \div hw^{OPs} \\
		sw^{GBs} \div hw^{GBs}
	\end{cases}
\end{equation}

Equation \ref{equation:roofline_time} can be rearranged to compute the actual~\emph{OPs} rate as described in Equation~\ref{equation:roofline_gbps}, that is, the minimum between the peak throughput of the hardware or what it can provide given the algorithms arithmetic intensity and the available HW memory bandwidth.
\begin{equation}
\small
	\label{equation:roofline_gbps}
	OPs = min
	\begin{cases}
		hw^{OPs} \\
		sw^{Arithmetic\ Intensity} \times hw^{GBs}
	\end{cases}
\end{equation}

\begin{figure}
	\centering
	\includegraphics[width=0.30\textwidth, page=2, trim={4.85cm 5.75cm 19.7cm 10.3cm},clip]{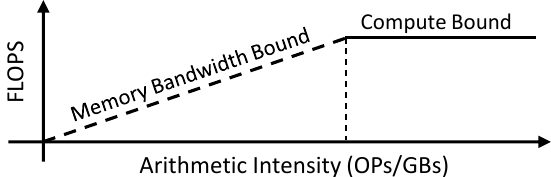} 
	\vspace{-2mm}
	\caption{Log scale plot of Equation \ref{equation:roofline_gbps}.}
	\label{figure:roofline_gbps_plot}
	\vspace{-5mm}
\end{figure}

By visualizing Equation~\ref{equation:roofline_gbps} in a log scale plot, as illustrated in Figure~\ref{figure:roofline_gbps_plot}, we may analyze the maximum throughput the hardware can provide for a given algorithm and input.
The horizontal line represents the peak \emph{OPs} of the hardware. The diagonal dashed line defines a linear relationship of arithmetic intensity and the peak bandwidth the hardware can provide. Algorithms and kernels can be profiled and then plotted into the locality plane. The Kernels close to the line are either bandwidth or compute bound and achieve good utilization of the HW resources. The Kernels distant from the line are prominent candidates for optimization.

In addition to the compute- and memory-bound workloads, tiny kernels are latency-bound, i.e.,  execution time is mainly derived from the overheads of launching work on the GPU. For simplicity, we have removed this case from Equation~\ref{equation:roofline_gbps}.

\section{DNNs - Training and Inference}
\label{sec:pipeline}

\begin{figure*}[!ht]
	\centering
	\vspace{3mm}
	\includegraphics[width=1\textwidth, page=3, trim={0.6cm 11.7cm 1cm 0.15cm},clip]{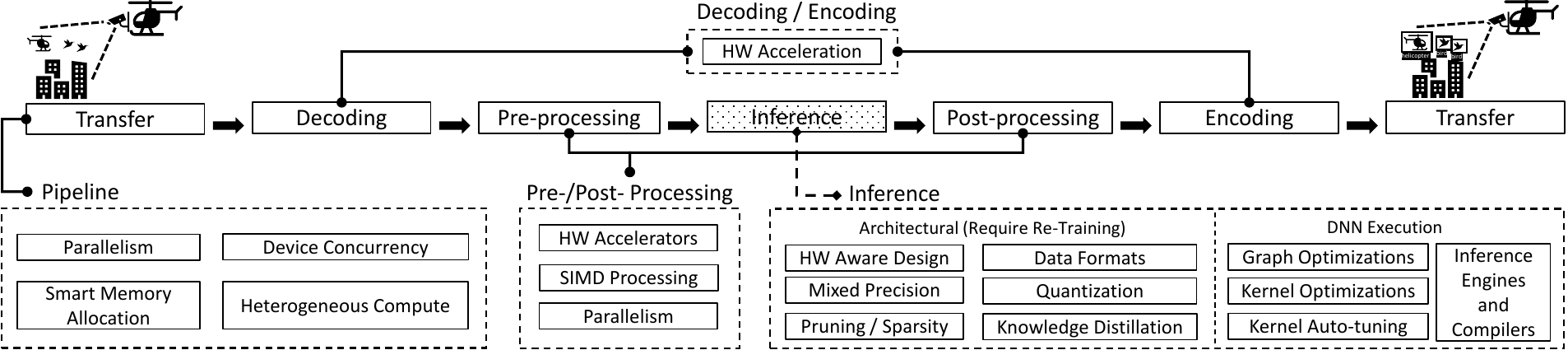} 
	\vspace{-6mm}
	\caption{ML processing pipeline for object detection and tracking. Each processing stage is characterized with the most prominent performance optimizations.}
	\label{figure:pipeline}
	\vspace{-4mm}
\end{figure*}

Among the most popular types of DNNs \cite{paper:history_began_from_alexnet} are Convolutional Neural Networks~(CNNs), Recurrent Neural Networks~(RNNs), and Transformers~\cite{paper:google_attention}. 
DNNs are composed of operators (layers) that perform well-defined mathematical operations on a set of features (data). Such computational style is often modeled as a directed computation graph where input/intermediate features and model parameters are represented as Tensors, a generalization of Matrices for dimensions larger than 2.

Alexnet~\cite{paper:alexnet} is composed of few pooled convolutional blocks, followed by 3~fully connected layers. In a modern implementation of Alexnet, the convolutional blocks would use a ReLU activation function, a Local Response Normalization block and a Max Pool layer. The fully collected blocks would also utilize an activation function such as ReLU or Softmax.
CNNs usually follow the same architectural patterns as Alexnet, where they receive an image of a given dimension with 3~RGB channels, and perform repeated convolutions with increasing channel count and decreasing dimensions composed of features representing higher level concepts.

The key differentiators that impact the accuracy of a DNN are the training dataset, the DNN architecture, and the training recipe~\cite{paper:timm}. Modern training datasets often include millions of classified samples~\cite{paper:imagenet}. 
The quickest way to designing a DNN model is to apply transfer learning~\cite{paper:resnext_wsl}, where an existing DNN that has been designed and trained for a similar problem is retrained and specialized to a new dataset or task. 
However, for specialized tasks, custom DNN architectures may bring significant accuracy and performance benefits. In a custom DNN, the architecture is tweaked by applying custom operators such as domain-specific activation functions, normalization layers, and other types of custom layers.
Finally, modern training recipes include techniques ranging from data augmentation to the DNN training procedure (loss function, optimizer, regularizer, hyper-parameters, and others), we refer to \cite{paper:timm} as an accurate survey of these techniques.

The process of developing DNNs can be characterized by two phases, referred as training and inference. During the training phase, a DNN model is designed and trained. During inference phase, a trained DNN model is integrated in an ML application and deployed on the target HW.  These two phases are similar and can be seen as a set of sequential operations (a pipeline), where the operations performed during training are a superset of the operations done during inference, i.e., data pre-processing during training may involve data augmentation.

In Figure~\ref{figure:pipeline}, we introduce a generic ML processing pipeline for image-based tasks. Such a processing pipeline is generally composed of data I/O (i.e, transferring images from a camera or drive), decode/encoding (i.e, decoding of compressed images), pre-/post-processing (i.e. data normalization/removal of redundant detections), DNN inference, and during training, back-propagation. 
During training or inference, the processing pipeline may become unbalanced and the stage with the longest execution time becomes a performance bottleneck. Ideally, the bottleneck should be the DNN execution as it is by far the most compute intensive task. In practice, DNN operators and inference engines are highly optimized, and thus, a different stage may also become a bottleneck.

Depending on the processing pipeline complexity and the computer architecture, some stages in the processing pipeline might be mapped to a CPU while others are mapped to a specialized HW accelerator. As an example, pre-/post- processing is often performed on a CPU, as it is easier to write CPU code for these phases, and allows computation to overlap with inference. As inference is computationally intensive, it is usually executed on a hardware accelerator such as a GPU.  

Pipeline performance bottlenecks generally fall within the following categories: (1) CPU-bound, which occurs when a GPU has low and spurious utilization rate as the CPU cannot keep the GPU saturated; (2) Synchronization-bound, which occurs whenever there are regular small-grained synchronizations (data-transfers) between the CPU and GPU, and both have low utilization; (3) GPU-bound, which occurs whenever the GPU has high utilization ($>$90\%) and is relatively stable without spikes or drops.  As performance implications may vary across ML applications and implementations,
a profiler is required to identify the bottlenecks. 

Figure \ref{figure:pipeline} suggests the most prominent performance optimization techniques for each stage of the processing pipeline. The listed optimization techniques alleviate bottlenecks from a given stage and thus increase overall performance. Decoding and encoding can be accelerated using specialized HW accelerators. Pre-/post- processing can be accelerated by leveraging SIMD processing and multi-core processing (parallelism).  For high-throughput, or whenever pre-/post processing is too computationally intensive, it might be necessary to perform these stages on the accelerator (e.g., FPGA and GPU).

DNN optimizations can be categorized into two groups referred as architectural and execution optimizations. The architectural optimizations re-design a DNN model and require re-training or calibration, while the execution optimizations preserve the original DNN model. The architectural optimizations include, (1) hardware-aware designs, which adapt the model architecture to fit a particular accelerator; (2) data formats, mixed-precision, and quantization, which use lower-precision formats to lower memory bandwidth requirements and leverage faster HW functional units; (3) pruning and sparsity, which reduce DNN computational intensity by removing neurons from a DNN; and (4) knowledge distillation, which aims to produce a smaller model from a larger parent. The execution optimizations include, (1) graph optimizations, such as layer-fusion which attempts to combine multiple operators to increase data-reuse (arithmetic intensity) and reduce launch overheads; (2) kernel optimizations, which are performance-optimized implementations of DNN operators that often leverage HW-dependent capabilities; (3) kernel auto-tuning, which attempts to derive which parameters and algorithms are fastest for a given operator and input dimensions; and (4) inference engines and compilers which often offer out-of-the-box support for all these techniques.

Finally, we list optimizations for the pipeline itself.  A smart memory allocation strategy that minimizes latency can often reduce or even eliminate synchronization-bottlenecks. In cases where high-throughput and low-latency are required and the computation within a phase is too small to fully leverage the dedicated accelerator, using device concurrency can reduce latency by performing multiple requests (frames) in parallel.

In the next sections, we describe DNN (Section~\ref{sec:nn_archs}) and pipeline (Section~\ref{sec:pipeline_optimizations}) optimizations in greater detail, including methods and tools proposed by both academia and industry.

\section{DNN Performance Optimizations}
\label{sec:nn_archs}

\subsection{DNN Operator Performance Analysis}

DNN operators can be classified within the following three categories~\cite{paper:nvidia_fp16_mixed_precision_training, site:nvidia_deep_learning_performance}, (1) Vector dot-product operations, such as Fully Connected Layers and Convolutions, which can be represented as matrix-vector or matrix-matrix multiplies; (2) Reduction operations, such as normalizations (batchnorm, etc.) and poolings (max-pool, etc.), which produce a series of values computed over a range of input tensors; (3) Point-wise operations that perform element-wise mathematical operations on the input tensor, such as the ReLU activation function, which performs $max(0,x)$ for all elements within a tensor.  

\begin{table}
	\footnotesize
	\centering
	\begin{tabular}{lc}
		\hline
		Operation & Arithmetic Intensity \\
		\hline
		Linear (4096 out, 1024 in, 512 b) & 315 FLOPS/B  \\[0ex]
		Linear (4096 out, 1024 in, 1 b) & 1 FLOP/B  \\[0ex]
		3x3 Max Pool, stride 1 & 2.25 FLOPS/B   \\[0ex]
		ReLU Activation & 0.25 FLOPS/B    \\[0ex]
		Layer Norm & $<$10 FLOPS/B   \\
		\hline
	\end{tabular} 
	\vspace{2mm}
	\caption{Arithmetic Intensity for common DNNs operators.}
	\label{table:nns_arithmetic_intensity}
	\vspace{-9mm}
\end{table}

Table~\ref{table:nns_arithmetic_intensity} provides a listing of the typical DNNs operators and their arithmetic intensities~\cite{site:nvidia_deep_learning_performance}. Using arithmetic intensity, the listed operations can be categorized on how efficiently the HW accelerator is leveraged.  Most operations have relatively low arithmetic intensity, sometimes even only performing a few operations per byte. 
Thus, out of the listed operations, only the first linear (fully connected) layer with a batch size of 512 can be classified as a compute-bound on a modern device. These properties depend on the compute capacity and memory bandwidth of a specific device, thus layers within a DNN should be profiled with the goal of finding the execution time breakdown per layer, and whether they are compute or memory bound.

Operations not represented as dot-products are almost always memory-bound. Only dot-products with sufficiently large input dimensions can be compute-bound. If an operation is memory-bound, then its execution time is dominated by the time required to move the necessary data and increasing computation speed will not improve performance. 

GPUs are inherently throughput-oriented devices,  and thus, only highly data-parallel and compute-intensive algorithms can fully occupy their computation resources. For DNNs, high-dimensioned input tensors (batch size, input and output size, channel counts, etc.) are required to saturate the compute and memory resources of a particular device. In addition, dimensions that facilitate the process of dividing the input matrix into blocks such that: (1) blocks have an even amount of work, and (2) blocks are evenly distributed and divisible among the many cores of a GPU, will lead to the best usage of GPU resources by preventing cores from remaining idle during execution.

\subsection{Layer Optimizations}
\label{sec:nn_optimizations}

\subsubsection{Data-formats and Quantization}
	While FP32 was the dominant format for ML applications, a variety of higher performance and lower precision formats such as TensorFloat32 (TF32)~\cite{paper:nvidia_a100}, IEEE FP16 mixed-precision training~\cite{paper:nvidia_fp16_mixed_precision_training}, and BFloat16 (BF16)~\cite{paper:bfloat16} have been proposed. These formats combine the range (8bit exponent) of a 32-bit floating point with the storage space of FP16 by reducing the mantissa, which is 10bits for TF32 and 7bits for BF16. As the support for these novel formats increases, they are becoming the new standard format to train DNNs~\cite{paper:timm}. 
	Once trained, DNNs are often deployed in even lower precision formats such as INT8~\cite{paper:nvidia_quantization}.

	The performance benefits of lower-precision formats is derived from the higher efficiencies of modern hardware. 
	While the extra compute capability is capable of accelerating compute-bound layers, it does not impact the memory-bound layers (e.g., batch-norm and neuron activations). For memory-bound layers, lower precision formats increase the performance by reducing memory bandwidth. As an example,  INT8 reduces FP32 memory bandwidth requirements by 4x. 
	INT8 not only requires less memory bandwidth, but also reduces chip size and power consumption as integer functional units are simpler in design. 

	For training, TF32 is a drop-in replacement for FP32, and BF16 comes close to being a drop-in replacement, requiring mixed-precision training. In FP16 mixed-precision training~\cite{paper:nvidia_fp16_mixed_precision_training}, some matrix operators are scaled in magnitude (i.e., multiplied by a constant) to keep them within the smaller range of FP16 without vanishing (becoming zero) or exploding (becoming infinite) and later accumulated in FP32. This is possible because DNN training uses a fraction of the original FP16 representable range for weights and activations. Thus, mixed-precision training preserves the accuracy of the FP32 training while performing some operations in FP16.

	Using INT8 precision is significantly more challenging and is only used for inference through quantization. Quantization is the process of translating a high-precision format (FP32) to lower a precision format (INT8), such that after performing some computation $C$, the weights may be de-quantized to their original precision and have the same values as if the computation $C$ was performed in its original high-precision. 

	While reducing the precision of a DNN from FP32 to FP16 is trivial for inference, INT8 quantization is challenging as the input space of INT8 is significantly smaller than FP16. 
	Thus, efficient scaling and compression techniques are required.

	The two most popular quantization techniques are: post-training quantization and quantization aware-training~\cite{paper:nvidia_quantization}. Post-training quantization initiates with a pre-trained DNN and uses a sample dataset to calibrate the dynamic ranges of the quantization parameters. In quantization-aware training, the parameters are learned during training; generally a pre-trained model is applied as a starting point and continuously tuned. Post-training quantization is usually fast while quantization-aware training is usually slow but allows fine-grained control over accuracy. State of the art quantization techniques can quantize DNNs with an accuracy loss within 1\% of the FP32 baseline across multiple application domains such as vision, speech, and language processing~\cite{paper:nvidia_quantization}. 
	For some DNN models with specific activation functions (e.g., Swish and Mish in Depthwise convolutions of MobileNet), the quantization results in high accuracy losses and state-of-the-art methods are required to maintain accuracy~\cite{paper:hard_quantization, paper:mobilenetv3}.
	 
	Although using quantization may reduce the accuracy of the original model, the increased performance creates opportunities to increase accuracy by: (1) using better/deeper/wider models with higher compute  requirements, (2) using higher resolution inputs, and (3) increasing the overall frame-rate.

\subsubsection{Pruning and Sparsity} 
	Pruning reduces the complexity of DNN architectures by removing neurons which have limited impact on the outcome, thus improving performance~\cite{paper:lecun_pruning}. State-of-the-art pruning methods have demonstrated significant reduction ($\sim$13x) of a model without impacting accuracy~\cite{han2015learning}. 
	
	Convolutional layers are form of pruning since the cross-product of each output feature only considers a small region within the input features. This leverages the compositional nature of the universe where objects are hierarchically composed. For example, sentences are composed of words which in turn are composed of letters. 
	Convolutions layers are inspired by the brain, which has been shown to be sparsely activated in a fashion similar to convolutional / pooling layers~\cite{thorpe2001seeking}.

	However, even convolutional layers can be pruned, since not all pixels within a given patch may be relevant. Selecting which neurons to prune across an entire DNN is an active area of research with two main directions~\cite{paper:pruning}. The first direction evaluates the contribution of each neuron in the network to the overall accuracy. Within a DNN, there are millions if not billions of neurons, thus these methods resort to analytical methods for efficiency. The second direction omits neurons that have the lowest weights in magnitude, as these are the least significant. 

	However, applying pruning to accelerate computation and reducing memory requirements is in practice a non-trivial problem. GPUs leverage highly efficient vector and matrix math pipelines that require dense matrices as input. Thus, omitting a neuron, or a group of randomly placed neurons while maintaining efficiency is a challenging problem. The performance benefits of pruning are often negligible and at times negative, even with high pruning rates as high as 95\%~\cite{paper:wen2016learning}.
	
	A form of Pruning is Sparsity, which aims to encourage zero values in parameters that can then be discarded~\cite{ paper:google_sparsity}. In structured Sparsity, 50\% of the model weights are removed by expunging 2~weights from each contiguous block of 4~weights~\cite{paper:nvidia_sparsity}. While structured sparsity is not able to reduce the original DNN model size as much as alternative methods, it synergizes well with hardware accelerators, and is thus able to achieve up to 2x speedup since 50\% of the computation is removed. 
	For DNN models such as Vision, Natural Language Processing, and Generative Adversarial Networks, no accuracy loss was observed with 2:4 structured sparsity~\cite{paper:nvidia_sparsity}.

	One question that may arise is, `if a DNN can be significantly pruned by removing a significant portion of its weights, wouldn't a smaller model achieve a similar result compared to the original DNN?' In practice, it is hard to train a smaller model and have it achieve the accuracy of the original. Frankle and Carbin~\cite{paper:nn_lottery_ticket_hypothesis} suggest that within a DNN model there is a smaller model that if trained on its own, would match the performance of its parent. However, determining the smaller model alone is challenging. In addition, the authors suggest that the random initialization applied in the DNN models are alike lottery tickets, increasing the chances that some initialization may assist the convergence during training. 	Finally, knowledge distillation aims to produce a more
	accurate model by distilling the knowledge of a larger model~\cite{paper:cai2020onceforall}; where the newly generated model may surpass the siblings trained in isolation.

\subsection{Graph Optimizations}
\label{sec:graph_optimizations}
	In this section we discuss graph optimizations that preserve the original DNN architecture.

	\subsubsection{Input Layouts}
		Frameworks, inference engines, and exchange formats employ different input layouts. The two most popular input layouts are channels-last (batch, height, width, channel) referred to as NHWC, and channels-first (batch, channel, height, width) referred to as NCHW. 
		Compute devices may have a preferred format, for Nvidia, channels-last offers the best performance since  Tensor Cores require this format. 

	\subsubsection{Layer Fusion, Reduction, and Elision}
		DNNs are composed of many layers, each performing a specific operation. 
		The motivation behind layer fusion is to increase efficiency by merging multiple operators together. In Alexnet, convolutional blocks are composed of a Convolution, a ReLU Activation, a Local Response Normalization, and a Max Pool. Using layer fusion, it is possible to combine these kernels into a single one. Such a kernel not only reduces overhead by minimizing the total number of kernel calls, but also opens the possibility for further optimizations by attempting to perform all of them together. Fusing compute-bound with memory bound kernels may increase  arithmetic intensity by fetching the convolution weights and input data from memory only once while keeping intermediate results in the cache.	

		Besides layer fusion, layer reduction and elision also improve arithmetic intensity, and thus increase performance. Layer reduction merges two operations into one, such as removing the bias of a convolution that is followed by a batch normalization that would subtract the mean and thus cancel out the bias. Layer elision removes concatenation operations by having the kernels directly write to the target tensor memory location.

		Layer fusion, reduction, and elision are dependent on the DNN architectures and require custom implementations. Inference engines such as Nvidia TensorRT and Intel OpenVino  provide support for many of the optimizations out of the box.

	\subsection{Kernel Optimizations}
	\label{sec:kernel_optimizations}
		DNNs are composed of a wide range of mathematical operators such as convolutions, normalizations, activations, and poolings which are implemented as kernels. Thus, optimizing these kernels for a particular HW and input dimensions can increase the performance of a wide range of models.  Kernel optimizations are inherently HW-dependent; as such we only provide a few examples of the optimizations, and recommend to refer to the HW vendors on how to fully leverage their platforms. 

		Exploiting cache locality is one of the most common optimizations to consider when optimizing GPU kernels. 
		A variation of cache locality that can be applied to memory-bound layers is persistency, where data is kept on the accelerator lower cache levels during execution. 

		Finally, we introduce the concept of auto-tuning which aims to select the best algorithms and parameters based on the target accelerator and input dimensions. For example, for matrix multiplication, there may be multiple algorithms available like GEMM or transform-based methods such as the FFT or Winograd algorithms. Each algorithm offers unique performance characteristics with respect to the input dimensions, batch size, tensor layout, etc. 
		TensorRT is an example of a tool which performs significant auto-tuning.
		TensorRT benchmarks multiple kernels, fusion options, kernel parameters, and data-formats to find the optimal kernel for a given operation. 

\section{Other Performance Optimizations}
\label{sec:pipeline_optimizations}
In this section, we introduce performance optimizations for the remaining processing stages in the ML application pipeline.

\subsection{Decoding and Encoding}
    For video and image processing, decoding is among the very first steps in the processing pipeline. Moreover, once the main processing has completed, an annotated video/image file often has to be created, thus requiring the data to be encoded. 
To alleviate potential encoding/decoding bottlenecks, some chip vendors have included HW acceleration for decoding/encoding. 

\subsection{Pre-processing and Post-Processing}
    Pre-processing covers data normalization, resizing, cropping, and others. In addition to pre-processing, some DNNs may require post-processing.
    At the edge, CPUs may not have sufficient computation capacity to pre-process data at a rate which saturates the GPU. With the help of SW libraries (e.g., Pillow-SIMD~\cite{site:pillow-simd}), pre-processing can be mapped to SIMD accelerators in the CPU or dedicated HW accelerators. For example, Xilinx provides a SW library and a specialized IP core~\cite{site:vitis_preprocessing} to accelerate pre/post processing. DALI, by Nvidia, provides  mixed GPU/CPU acceleration~\cite{site:nvidia_dali}. Finally, Albumentations~\cite{paper:albumentations} and Kornia \cite{paper:kornia} support various HW options and deliver wide range of pre-processing operators for audio, image, and video.

\subsection{Processing Pipeline}

    \subsubsection{Parallelism}
        Chip vendors have started to provide tools and examples that leverage the hardware to its full potential. Nvidia has released DeepstreamSDK 
        that interfaces with GStreamer and extends it with custom set of plugins. Xilinx provides a similar approach with the Video Analytics SDK.

    \subsubsection{Device concurrency}
        With small DNN models, it might become a challenge to saturate a modern GPU. Increasing the batch size can increase GPU utilization at the cost of higher latency. Thus, there is a trade-off between throughput and latency that must be considered. 
        A way to increase the performance without significantly impacting latency is to dispatch multiple inference requests in parallel on a device. 

    \subsubsection{Smart shared-memory allocation}
        The smart shared-memory allocation allows large transactions to be performed in batches that may overlap with other operations such as data computations. Ideally, the memory shall be pre-allocated and re-used on both the HW accelerator and the CPU to avoid allocation overheads at runtime. Data transfers shall be minimized.

    \subsubsection{Heterogeneous Compute}

        A way to improve performance is to use multiple HW accelerators to infer a single DNN. For example, (1) to distribute the network among several HW accelerators and achieve higher frame-rate or lower latency, (2) to run incompatible or hard-to-implement operators on a more developer friendly device such as CPU. However, distributing the DNN across multiple HW accelerators needs to be performed with care as it is easy to lose performance. Compute-bound layers with high arithmetic intensity (e.g., convolutions) should execute on a HW accelerator. Nevertheless, the performance benefits of the heterogeneous execution heavily depends on the communication granularity between the available devices. 
		If the data transmission/ conversion is longer than the execution itself, then there is no performance benefit in using multiple HW accelerators. 
		For example, CPUs in particular do not support FP16, thus mixing FP16 (GPU) and FP32 (CPU) can result in conversions that lead to performance bottlenecks.

\section{Experimental Section}
\label{sec:experimental}

In this section we present a practical example on how to analyze, optimize, and deploy a state-of-the-art DNN. As a DNN of choice, we select Recurrent All-Pairs Field Transforms (RAFT)~\cite{paper:RAFT}, an advanced deep network architecture for optical flow. 
As an edge AI inference platform, we chose the Nvidia Jetson AGX Xavier. 

RAFT has a total of 5.26~mil parameters. With an input resolution of 1224x370, RAFT requires $\sim$1.1 TFLOPS during forward propagation. With the Jetson's GPU computational resources (10 TFLOPS in FP16/ 20 TOPS in INT8), the performance upper-bound is $\sim$19/10 FPS in INT8/FP16. This analysis assumes peak HW throughput; in practice, best-case performance is anticipated to be significantly lower.

To optimize inference performance, we export RAFT from Pytorch to TensorRT using ONNX. We chose TensorRT as it is performance-optimized for Nvidia devices, and includes many of the performance optimizations referenced in previous sections. Since RAFT uses the grid sampler operator, which is not part of ONNX opset 13 nor supported by TensorRT 8.2.0, we implement a custom operator via a TensorRT plugin.

\begin{figure}
	\centering
	\includegraphics[width=0.499\textwidth, page=1, trim={2cm 6.95cm 1.2cm 7cm}, clip]{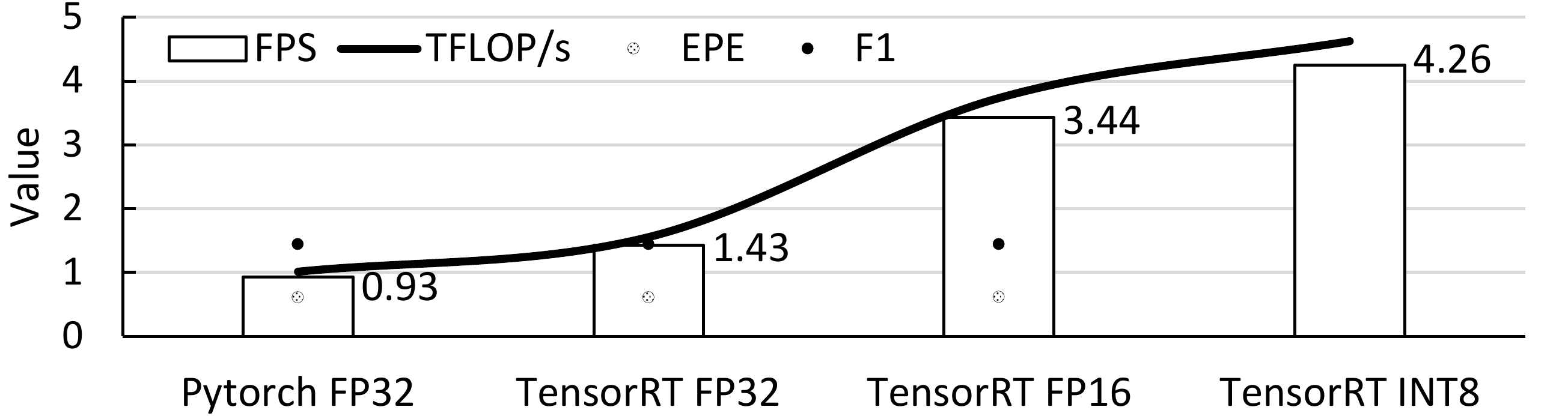}
	\vspace{-6mm}
	\caption{Performance and Accuracy Metrics for the RAFT DNN.}
	\label{figure:perf_raft} 
	\vspace{-5.7mm}
\end{figure}

In Figure~\ref{figure:perf_raft}, we introduce the performance and accuracy metrics for both the Pytorch and the TensorRT designs. The original Pytorch design achieves  0.95~FPS; profiling reveals that the low performance is due to the Pytorch (Python) run-time which becomes a bottleneck on the Jetson as it is unable to saturate the GPU. 
The TensorRT runtime removes the bottleneck and shows higher performance at 1.43~FPS. Using FP16 inference delivers a 2.4x performance boost without impacting accuracy. Additional performance can be achieved by quantizing RAFT from FP16 to INT8. 
However, we observed total accuracy loss during INT8 quantization. 
We speculate that the numerical distribution within RAFT is too broad to be rescaled and compressed within INT8. Quantization-aware training and architectural changes would be necessary to encourage numerical ranges that are quantizable. Examples of such transformations include adding batchnorm layers or activations functions which normalize/bind the activation values within a given numerical range. Given the highly specialized nature of RAFT, not all computational blocks within RAFT may be quantizable, and partial FP16 execution may be required.
 
At a performance ceiling of 10~FPS in FP16 mode, the achieved 3.44~FPS seems somewhat low. 
The biggest bottleneck is due to the custom grid sampler operator, which requires tensors in channels-first format, while the Tensor Cores operate best in channels-last format. 
To estimate the total performance benefit of 
channels-last inference, we replaced the grid sample operator by a no-op. 
Using this method, we obtained more than half of the peak GPU performance in FP16 at 6.13~TFLOPS out of 11~TFLOPS.

Although we have boosted performance up to 6x in FP16, further optimizations are possible. 
First, in a series of sequential images, the second image input of inference $N$ becomes the first input of inference $N+1$, thus it is possible to re-utilize computation. Second, the Jetson DLAs can be utilized to off-load some computation. Third, modifications to the DNN model might be applicable, such as pruning/sparsity and knowledge distillation.

\section{Conclusion}
\label{sec:conclusion}

In this paper, we provided an overview of DNNs and specify their execution and performance profiles. We reviewed the state-of-the-art software and hardware optimizations that improve the performance of DNNs while preserving accuracy. These optimizations ranged from pipeline optimizations, that enhance the performance of the computation surrounding the DNN, such as encoding/decoding and pre-/post-processing, to optimizations of the DNN itself.  We applied a set of optimizations to a state-of-the-art DNN and reduced execution time significantly. 

\section*{Acknowledgment}
Supported by the European Union’s Horizon 2020 research and innovation programme under grant agreements No. 945535.

\bibliographystyle{IEEEtran}
\bibliography{paper}

\end{document}